\pdfoutput=1

\documentclass[11pt]{article}

\usepackage{acl}

\usepackage{times}
\usepackage{latexsym}
\usepackage{booktabs} 
\usepackage{graphicx}
\usepackage{bbding}
\usepackage{multirow}
\usepackage{subcaption}
\usepackage{tablefootnote}

\usepackage[T1]{fontenc}

\usepackage[utf8]{inputenc}

\usepackage{microtype}

\newcommand{\largemodel}{PEGASUS-X}
\newcommand{\smallmodel}{{PEGASUS-X\textsubscript{Base}}}
\newcommand{\pegasuslarge}{PEGASUS\textsubscript{Large}}
\newcommand{\pegasusbase}{PEGASUS\textsubscript{Base}}
\newcommand{\pegasusbaseplus}{PEGASUS\textsubscript{Base+}}
\newcommand{\pegasusbaseplusgl}{PEGASUS\textsubscript{Base+} +  Global-Local}
\newcommand{\rougecolumns}{c@{/}c@{/}c@{\hspace{\tabcolsep}}c}
\newcommand{\rougecolumnnames}{& \hspace{2pt} R1 \hspace{2pt} &  \hspace{2pt} R2  \hspace{2pt} &  \hspace{2pt} RL \hspace{2pt} & RG}
\newcommand{\rougecolumnnameslsum}{& \hspace{2pt} R1 \hspace{2pt} &  \hspace{2pt} R2  \hspace{2pt} &  \hspace{2pt} RLs \hspace{2pt} & RG}
\title{Investigating Efficiently Extending Transformers \\ for Long Input Summarization}

\author{
Jason Phang \textsuperscript{1}\thanks{\hspace{2mm}Work done while at Google.}~~~~~~Yao Zhao\textsuperscript{2}~~~~~~Peter J. Liu\textsuperscript{2}\\
\textsuperscript{1}New York University,~~
\textsuperscript{2}Google Research, Brain Team\\
{\tt jasonphang@nyu.edu} \\
{\tt \{yaozhaoyz, peterjliu\}@google.com}\\
}

\begin{document}
\maketitle
\begin{abstract}
While large pretrained Transformer models have proven highly capable at tackling natural language tasks, handling long sequence inputs continues to be a significant challenge.
One such task is long input summarization, where inputs are longer than the maximum input context of most pretrained models.
Through an extensive set of experiments, we investigate what model architectural changes and pretraining paradigms can most efficiently adapt a pretrained Transformer for long input summarization.
We find that a staggered, block-local Transformer with global encoder tokens strikes a good balance of performance and efficiency, and that an additional pretraining phase on long sequences meaningfully improves downstream summarization performance.
Based on our findings, we introduce \largemodel{}, an extension of the PEGASUS model with additional long input pretraining to handle inputs of up to 16K tokens.
\largemodel{} achieves strong performance on long input summarization tasks comparable with much larger models
while adding few additional parameters and not requiring model parallelism to train.
\end{abstract}


\section{Introduction}

Large pretrained Transformer models have proven to be extremely capable at tackling natural language tasks \citep{devlin2018bert,brown2020gpt3}. However, handling long textual sequences continues to be a significant challenge for these models.
Training models to handle long sequences is expensive in both computation and memory, and moreover requires training and evaluating on long sequence data, which can be rarer and more costly to collect.
Given the broad success of Transformer models on short-sequence language tasks, our goal is to investigate the best way to extend these models to handle longer sequences.

\begin{figure}
  \centering
  \includegraphics[width=\linewidth]{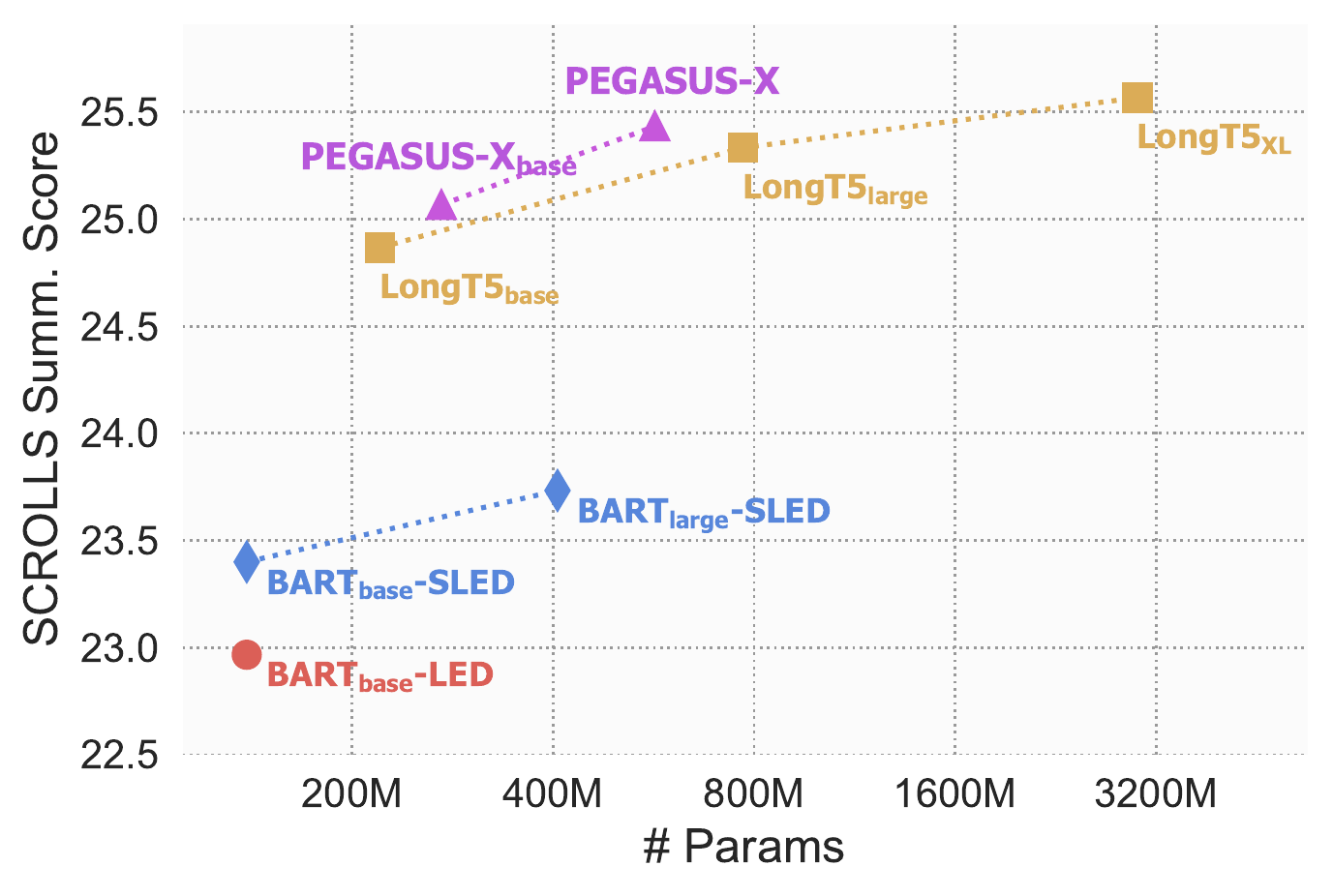}
  \caption{
    Model scores on SCROLLS \citep{shaham2022scrolls} summarization tasks.
    All models evaluated on up to 16K input tokens.
    PEGASUS-X outperforms other models at comparable model sizes.
    Scores are computed by taking the average of the geometric mean of ROUGE-1/2/L.
  }
\label{fig:summary_plot}
\end{figure}

In this work, we focus on the task of long input summarization: summarizing long input documents into shorter textual sequences.
The input documents of such tasks are often significantly longer than the maximum context lengths of most standard Transformer models, and hence warrant both specialized model architecture modifications as well as new training regimes to handle.
For instance, to avoid the quadratic growth in memory consumption of the attention computation in Transformers, many memory-efficient Transformer variants have been proposed \citep{tay2020efficient,tay2021long}.
However, the manner in which these changes are incorporated into models has been inconsistent and ad-hoc, and there are few established best-practices.
For instance, some works add an additional long input pretraining stage to adapt the model weights to the new architecture \citep{beltagy2020longformer}, while others directly fine-tune on the long-input summarization data without any pre-adaptation \citep{zaheer2020bigbird, pang2022topdown}.
Because of the high cost of training these models, there has yet to be a systematic study of how best to adapt models for long input sequences.
Hence, it has been difficult to establish which model and training changes are necessary or complementary.

To answer these questions, we conduct an extensive empirical investigation into the architectural changes, model configurations and pretraining schemes to identify the better approaches to training Transformer models to tackle long input summarization.
We evaluate a set of efficient Transformer variants, and propose a simpler block-wise local Transformer architecture with staggered blocks and global tokens that strikes a good balance of performance and memory efficiency.
We also show that given a fixed token budget, pretraining on short sequences and then pre-adapting the model to an efficient Transformer architecture on long sequence for additional training steps leads to superior performance compared to only long input pretraining or no adaptation at all.
We also investigate several other model design choices such as position encoding schemes, encoder-decoder layer distributions, and the impact of discrepancies between pretraining and fine-tuning architecture hyperparameters.

Based on the findings from our empirical investigation, we adapt the pretrained PEGASUS\textsubscript{Large} model \citep{zhang2019pegasus} to tackle long input summarization on up to 16K input tokens. 
The resulting model, which we call \largemodel{} attains top scores on long summarization tasks, outperforming much larger models like LongT5 \citep{guo2021longt5} in some cases, and sets the state of the art of two tasks: GovReport and PubMed.
Morever, impact on short input summarization performance is minimal.
A smaller version which we call \smallmodel{} attains similar scores with much fewer parameters.
The code and weights for both models will be released at \url{https://github.com/google-research/pegasus} and as well as in Hugging Face Transformers \citep{wolf2020transformers}.
Beyond long input summarization, we believe that many of our findings will be useful to the community for efficiently adapting Transformer models to handle ever longer input sequences for other tasks.

In summary, our contributions are:

\begin{enumerate}
    \item We evaluate a series of proposed efficient Transformer architectures as well as a host of other model tweaks, and report their efficacy as well as trade-offs on computational resources when applied to long input summarization tasks.
    \item Based on our findings, we propose a recipe for adapting a short-context, pretrained Transformer encoder-decoder to longer inputs, and apply it to PEGASUS to greatly improve its long-document summarization performance, with comparable short-input performance. 
    \item We release model checkpoints for the resulting 568M-parameter model, which we call \largemodel{}, and a smaller 272M-parameter model with most of the performance, \smallmodel{}.
\end{enumerate}

\section{Challenges of Long Input Summarization}

\subsection{Computational Challenges}

While summarization is fundamentally about extracting and compressing information from longer to shorter sequences, most commonly studied summarization tasks have had inputs on average shorter than the input sequence lengths of Transformer language models--typically 512 to 2048 tokens.
As the ability for models to handle language has improved, the field has pushed for more challenging summarization tasks with longer input lengths.
The quadratic scaling of the memory requirements and computation for the attention mechanism in Transformers poses a challenge to tackling these longer summarization tasks.
Many memory- and compute-efficient variants of Transformers \citep{beltagy2020longformer,zaheer2020bigbird,choromanski2021performer,wang2020linformer,kitaev2020reformer} have been proposed to address this constraint.
However, even when incorporating efficient Transformer architectures that achieve approximately linear memory scaling with input sequences, it is still common for models to be pretrained on short sequence inputs and only be adapted to handle long sequences when fine-tuning on a downstream task, which may be suboptimal.

While using decoder-only autoregressive language models for summarization has received some recent attention \citep{radford2019gpt2,brown2020gpt3,chowdhery2022palm}, encoder-decoder models still generally perform better and remain the architecture of choice for the task \citep{wang2022language}.
The asymmetry between the input length and summary lengths requires new considerations for resource limitations of models.
Consider a summarization model with 12 encoder and 12 decoder layers, pretrained on an input length of 512 and fine-tuned on a task with input sequence length 16384, using output length of 512 in both cases.
Since pretraining is typically done with shorter sequences while fine-tuning uses long inputs for summaries, fine-tuning can now be more resource intensive and slower than pretraining, which is contrary to the conventional paradigm.
Since the encoder inputs have increased 32$\times$, the quadratic scaling in the memory consumption of the self-attention operation means that we expect the encoder self-attention to consume 1024$\times$ the amount of memory in fine-tuning relative to pretraining.
Even if we use an efficient Transformer variant that achieves linear scaling in memory consumption and computation, both the encoder self-attention and decoder cross-attention operations still consume 32$\times$ the memory compared to pretraining.
Besides attention, expensive operations such as the FFN that scale linearly with the input length also greatly increase the computation required both at training and inference.

On the other hand, the unique characteristics of long-document summarization may also prompt new solutions to these issues.
For instance, if encoder computations over long sequences pose a compute bottleneck, we may consider using fewer encoder layers and more decoder layers, exchanging decoding speed at inference for faster training.
The higher relative cost of fine-tuning can also justify greater efforts to adapt the pretrained model to fine-tune more quickly, via mixing short- and long input training curricula, adapting the model to efficient Transformer architectures via additional pretraining, and so on.

To address these questions and challenges, we conduct a series of ablation experiments investigating which approaches can lead to improvements in downstream summarization results, as well as the computational trade-offs therein.

\subsection{Task/Dataset Challenges}

A challenge in building long-document summarization models is the relative scarcity of long-input summarization datasets with sufficient data to train and evaluate models on.
Recent work introducing new long-document summarization datasets has alleviated this problem somewhat \citep{chen2022summscreen, shaham2022scrolls,kryscinski2021booksum}, although the relative scarcity of good datasets continue to make this a challenging problem to make progress on.
The main issues in current datasets are: relative simplicity of summarization, lack of diverse inputs, potential leakage of data due to the data collection procedure, and low quantity of examples for training.
We refer the reader to \citet{wang2022squality} for more discussion on the challenges of creating large, high-quality long-document summarization datasets.

\section{Experimental Setup}

Similar to \citet{zhang2019pegasus}, we perform the majority of our experiments with a \pegasusbase{}-sized model, before applying our findings to \pegasuslarge-sized model.

\subsection{Pretraining}

\label{sec:exp_pretraining}

We generally follow the recipe from PEGASUS \citep{zhang2019pegasus} for pretraining \pegasusbase{}-sized models.
All experiments in our ablation study performed pretraining with C4 \citep{raffel2020t5} for 500k steps with 512 input tokens and 256 output tokens and a masking ratio of 45\%, unless otherwise stated.
For long input pretraining we extend the input length to 4096 tokens, adjust the masking ratio from 45\% to 5.625\%, reducing the ratio by a factor of 8 to account for the 8x increase in input sequence length.
We also filter for only documents longer than 10000 characters.

\subsection{Fine-tuning}

\begin{table*}[th]
\centering
\small
\resizebox{\textwidth}{!}{%
\begin{tabular}{l \rougecolumns \rougecolumns \rougecolumns \rougecolumns  cc}
    \toprule
    & \multicolumn{4}{c}{XSUM} 
    & \multicolumn{4}{c}{CNN/DM}
    & \multicolumn{4}{c}{arXiv} 
    & \multicolumn{4}{c}{GovReport}
    \\
    \cmidrule(lr){2-5}\cmidrule(lr){6-9}\cmidrule(lr){10-13}\cmidrule(lr){14-17}
    \cmidrule(lr){18-18}\cmidrule(lr){19-19}
    Encoder
    \rougecolumnnames
    \rougecolumnnames
    \rougecolumnnames
    \rougecolumnnames
    & Steps/s & Mem
    \\ \midrule
    Transformer
        & \textbf{40.0} & \textbf{16.9} & \textbf{32.0} & \textbf{27.9}
        & \textbf{39.5} & \textbf{19.0} & \textbf{28.6} & \textbf{27.8}
        & - & - & - & - 
        & - & - & - & - 
        & -
        & -
    \\
    BigBird
        & 39.6 & 16.7 & 31.7 & 27.6
        & 39.3 & 18.2 & 28.1 & 27.2
        & 46.8 & 19.6 & 28.0 & 29.5
        & 60.5 & 28.5 & 30.1 & 37.3
        & 0.31
        & 1.88
    \\
    Performer
        & 36.5 & 14.0 & 28.7 & 24.5
        & 37.4 & 17.4 & 26.9 & 26.0
        & 39.0 & 13.2 & 23.8 & 23.1
        & 55.8 & 20.2 & 24.7 & 30.3
        & 0.96
        & 1.12
    \\
    \midrule
    Local
        & 38.5 & 15.7 & 30.6 & 26.4
        & 39.0 & 18.4 & 28.1 & 27.2
        & 46.5 & 19.7 & 27.9 & 29.5
        & 60.2 & 28.3 & 30.0 & 37.1
        & \textbf{1.00}
        & \textbf{1.00}
    \\
    Global-Local
        & 38.7 & 16.2 & 31.2 & 26.9
        & 39.0 & 18.6 & 28.2 & 27.3
        & \textbf{47.6} & \textbf{20.2} & \textbf{28.5} & \textbf{30.1}
        & \textbf{61.4} & \textbf{29.3} & \textbf{30.6} & \textbf{38.0}
        & 0.87
        & 1.08
    \\
    \bottomrule
\end{tabular}%
}
\caption{
  Comparison of different encoder architectures on short (XSUM, CNN/DM) and long (arXiv, GovReport) summarization tasks.
  Training steps per second and memory are computed based on arXiv, and normalized to Local Transformer performance.
}
\label{tab:table_1_encoder_baselines}
\end{table*}

We evaluate our pretrained models by fine-tuning on the arXiv \citep{cohan2018arxivpubmed} and GovReport \citep{huang2021govreport} long-context summarization tasks.
Where relevant, we also fine-tune on the shorter-context XSUM and CNN/DailyMail tasks.
For each experiment, we report the best validation set scores based on the geometric average (RG) of ROUGE-1, ROUGE-2 and ROUGE-L scores \citep{lin2004rouge} based on the \texttt{rouge-score} package.\footnote{\url{https://github.com/google-research/google-research/tree/master/rouge}}
For arXiv, we fine-tune with an input length of up to 16384 tokens and 256 output tokens, while for GovReport we use an input length of 10240 input tokens and 1024 output tokens given the longer summaries for the task.
For XSUM and CNN/Daily Mail, with use an input length of 512, and output lengths of 64 and 128 respectively, following PEGASUS hyperparameters.
The full set of hyperparameters for fine-tuning can be found in Appendix~\ref{appendix:finetuning_hp}.
Unless otherwise stated, we directly switch over to the efficient Transformer architectures between pretraining (on shorter context) and fine-tuning (on longer contexts), with no adaptation phase in between.

\section{Experiments}

\subsection{Encoder architectures}

\label{section:ablation_encoder}

\begin{table*}[th]
\centering
\small
\begin{tabular}{lcc \rougecolumns \rougecolumns}
    \toprule
    & \multirow{2}{*}{\shortstack{Stagger \\ Local Blocks}}
    & \multirow{2}{*}{\shortstack{Use Global \\ In Decoder}}
    & \multicolumn{4}{c}{arXiv} 
    & \multicolumn{4}{c}{GovReport}
    \\
    \cmidrule(lr){4-7}\cmidrule(lr){8-11}
    Encoder
    & &
    \rougecolumnnames
    \rougecolumnnames
    \\ \midrule
    Global-Local & {\scriptsize \Checkmark} & {\scriptsize \Checkmark} 
        & \textbf{48.1} & \textbf{20.3} & \textbf{28.5} & \textbf{30.3}
        & 60.5 & 28.8 & 30.5 & 37.6
    \\
    Global-Local &  & {\scriptsize \Checkmark} 
        & 47.0 & 19.5 & 27.9 & 29.5
        & 60.9 & 28.9 & 30.2 & 37.6
    \\
    Global-Local & {\scriptsize \Checkmark} & 
        & 47.7 & 20.4 & 28.6 & 30.3
        & \textbf{61.3} & \textbf{29.4} & \textbf{30.8} & \textbf{38.1}
    \\
    Global-Local &  & 
        & 46.7 & 19.5 & 27.9 & 29.4
        & 59.5 & 27.8 & 29.4 & 36.5
    \\
    Local & {\scriptsize \Checkmark} & -
        & 46.8 & 19.7 & 28.0 & 29.6
        & 59.2 & 27.9 & 30.0 & 36.7
    \\
    Local &  & -
        & 46.5 & 19.2 & 27.5 & 29.1
        & 58.8 & 27.5 & 28.9 & 36.0
    \\
    \bottomrule
\end{tabular}%
\caption{
  Comparison of architectural tweaks to Local and GlobalLocal encoder.
  Staggering local blocks uses different blocks boundaries for different layers in block-local attention.
  Global information is incorporated in the decoder via an additional cross-attention before cross-attention over the encoded input.
}
\label{tab:table_2_global_and_local_configs}
\end{table*}

\begin{figure*}
\begin{subfigure}{.5\textwidth}
  \centering
  \includegraphics[width=.6\linewidth]{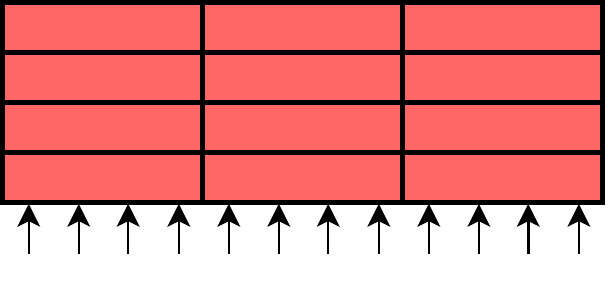}
  \caption{Block-local attention}
  \label{fig:staggering_a}
\end{subfigure}%
\begin{subfigure}{.5\textwidth}
  \centering
  \includegraphics[width=.6\linewidth]{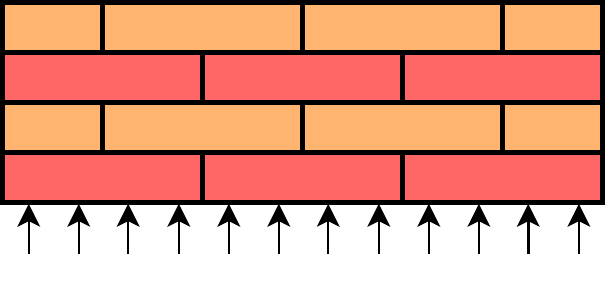}
  \caption{Block-local attention with staggered blocks}
  \label{fig:staggering_b}
\end{subfigure}
\caption{
  In block-local attention (a), the same block boundaries are used across all layers, preventing information from being shared across blocks.
  Staggering the block boundaries (b) be shifting the boundaries every other layer allows for cross-block interactions with minimal additional computational cost or complexity.
}
\label{fig:staggering}
\end{figure*}

We begin by investigating the efficacy of swapping the encoder for an efficient Transformer encoder to allow our models to incorporate longer input sequences while consuming reasonable amounts of device memory.
We first consider two efficient encoder architectures that exemplify two different approaches to memory-efficient attention.
Big Bird \citep{zaheer2020bigbird} takes the approach of using sparse attention computation, combining sliding-window attention, random attention and a set of global-attention tokens.
Conversely, Performer \citep{choromanski2021performer} takes the approach of factorizing attention matrices via orthogonal random features.
Both Big Bird and Performer have the benefit of requiring no new parameters to be introduced, and hence the weights from a pretrained Transformer can be ported directly to these architectures.
Both model also performed well on the Long Range Arena tasks \citep{tay2021long}.
However, for this experiment, we perform both pretraining and fine-tuning with the same encoder architecture to avoid the issue of mismatch between pretraining and fine-tuning architectures.

In addition, we also introduce two simple variants of local attention Transformer encoders.
First, we use a simple block-local Transformer (Local), where encoder input tokens are divided into non-overlapping blocks, tokens can only attend to other tokens within the block.
Second, we extend this local Transformer by adding a set of global tokens with learnable embeddings, that can attend to and be attended from every encoder token (Global-Local).
These components are similar in principle to the sliding window attention and global token attention of Big Bird, as well as similar constructs in other efficient Transformers such as ETC \citep{ainslie2020etc} and Longformer \citep{beltagy2020longformer}.
However, we opt for the simpler block-local attention rather than sliding window attention, and compensate for the lack of overlapping blocks by staggering the local attention blocks, which we elaborate on in Section~\ref{section:ablation_global_and_local_configs}.
As we show below, the performance is highly competitive despite its simplicity.

BigBird, Local and Global-Local all use a block size of 64, and 32 global tokens where relevant. Performer uses 256 random features.

Results on short and long summarization tasks are shown in Table~\ref{tab:table_1_encoder_baselines}, with the relative training steps per second and memory consumed per device for fine-tuning on arXiv shown in the right-most columns.
Among the short tasks, the full-attention Transformer performs best, followed by BigBird.
On the long tasks, Big Bird and Global-Local models perform best, but Big Bird consumes significantly more memory and trains much more slowly than the other architectures.
Conversely, we find that although the Performer has relatively low memory consumption and trains efficiently, it performs the worst out of the architectures we tested by a noticeable margin.

On the other hand, we find that the Local and Global-Local encoders strike a good balance of both performance and efficiency. 
The simple local attention encoder, which uses a block-local attention mechanism, attains performance surprisingly close to that of Big Bird while being much faster and using much less memory.
The Global-Local encoder trades off a small amount of speed and memory for better performance, outperforming Big Bird.
While both Local and Global-Local models underperform Big Bird and Transformer for short-tasks, it appears that the model architectures make the right trade-offs for performance on long summarization tasks.

\paragraph{Takeaways:} Local attention is a surprisingly strong baseline, while adding global tokens significantly improves performance, and both models are resource-efficient.


\subsection{Local and Global-Local configurations}

\label{section:ablation_global_and_local_configs}

Given the good performance of both Local and Global-Local encoder variants, we next consider further architectural tweaks to these models.

First, we introduce \emph{\textbf{staggering}} of local attention blocks. Unlike in sliding window attention, in block-local attention tokens can only attend to other tokens within the same block.
If the input tokens are divided up into the same blocks in every layer, this means that no information is exchanged across blocks through the entire encoder.
To address this pitfall, we introduce a small architectural change wherein we stagger the block allocation across alternating layers.
We show an example of this in Figure~\ref{fig:staggering}.
Concretely, we stagger attention blocks by shifting the block boundaries by half a block every other layer: in practice, we implement this by padding the hidden representations on either side by half a block and masking accordingly.

Secondly, in the Global-Local model, the decoder only attends to the encoded token representations, and not the global token representations.
We consider a variant where we supply the global token representations to the decoder, and in particular introduce a second encoder-decoder cross-attention that attends only to the global tokens, before performing cross-attention over the encoded tokens.
Our goal is to allow the decoder to incorporate global information before performing cross-attention over the encoded sequence.

We show the results of both of these changes in Table~\ref{tab:table_2_global_and_local_configs}.
We find that staggering local blocks improves performance in both Local and Global-Local models by a noticeable amount.
We highlight that this improves performance even with the Global-Local models, which already has a channel for cross-block interactions via global tokens, indicating that both of these model improvements are complementary.
Conversely, we did not find that incorporating global token information in the decoder led to much of a performance improvement, particular once staggered local blocks were used.

\paragraph{Takeaways:} Staggering local attention blocks significantly improves performance, and is complementary to global tokens.


\subsection{Global-Local: Block Size and Number of Global Tokens}

\begin{table*}[th]
\centering
\small
\begin{tabular}{rr \rougecolumns \rougecolumns cc}
    \toprule
    &
    & \multicolumn{4}{c}{arXiv} 
    & \multicolumn{4}{c}{GovReport}
    \\
    \cmidrule(lr){3-6}\cmidrule(lr){7-10}
    Block Size
    & Global Tokens
    \rougecolumnnames
    \rougecolumnnames
    & Steps/s & Mem
    \\ \midrule
    4 & 8
        & 46.2 & 19.1 & 27.5 & 29.0
        & 60.1 & 28.0 & 29.7 & 36.8
        & 0.77
        & 1.27
    \\
     & 32
        & 46.1 & 18.8 & 27.2 & 28.7
        & 60.1 & 27.6 & 28.9 & 36.3
        & 0.65
        & 1.70
    \\
     & 64
        & - & - & - & -
        & 60.1 & 27.7 & 29.0 & 36.4
        & - 
        & - 
    \\
     & 128
        & - & - & - & -
        & - & - & - & -
        & - 
        & - 
    \\ \midrule
    16 & 8
        & 46.9 & 19.6 & 27.9 & 29.5
        & 60.1 & 28.2 & 29.7 & 36.9
        & 0.98
        & 1.03
    \\
     & 32
        & 47.1 & 20.0 & 28.3 & 29.9
        & 59.7 & 27.8 & 29.2 & 36.5
        & 0.92
        & 1.15
    \\
     & 64
        & 46.8 & 19.7 & 28.0 & 29.6
        & 60.8 & 28.6 & 30.0 & 37.4
        & 0.75
        & 1.54
    \\
     & 128
        & 47.7 & 20.0 & 28.2 & 30.0
        & 60.7 & 28.8 & 30.2 & 37.5
        & 0.58
        & 1.70
    \\ \midrule
    64 & 8
        & 46.8 & 19.8 & 28.0 & 29.6
        & 61.2 & 28.8 & 30.2 & 37.6
        & 0.98
        & 1.06
    \\
     & 32
        & 47.7 & 20.3 & 28.5 & 30.2
        & 61.0 & 29.3 & 30.8 & 38.0
        & 0.47
        & 1.07
    \\
     & 64
        & 47.4 & 20.2 & 28.5 & 30.1
        & 60.9 & 29.1 & 30.7 & 37.9
        & 0.94
        & 1.10
    \\
     & 128
        & \textbf{47.8} & \textbf{20.4} & \textbf{28.6} & \textbf{30.3}
        & 60.9 & 29.0 & 30.3 & 37.7
        & 0.85
        & 1.26
    \\ \midrule
    128 & 8
        & - & - & - & -
        & - & - & - & -
        & - 
        & - 
    \\
     & 32
        & 46.9 & 19.7 & 28.0 & 29.6
        & 60.9 & 28.7 & 30.1 & 37.5
        & \textbf{1.00}
        & \textbf{1.00}
    \\
     & 64
        & 47.4 & 20.2 & 28.4 & 30.1
        & 60.9 & 28.9 & 30.8 & 37.8
        & 0.96
        & 1.05
    \\
     & 128
        & 47.1 & 20.0 & 28.3 & 29.9
        & 61.0 & 28.9 & 30.6 & 37.8
        & 0.90 
        & 1.15
    \\ \midrule
    256 & 8
        & 46.8 & 20.0 & 28.2 & 29.8
        & 60.7 & 29.3 & 30.9 & 38.0
        & 0.96
        & 1.07
    \\
     & 32
        & 47.3 & 20.2 & 28.3 & 30.0
        & 61.6 & 29.4 & 30.7 & 38.2
        & 0.92
        & 1.11
    \\
     & 64
        & 47.2 & 20.2 & 28.4 & 30.0
        & 59.2 & 28.6 & 30.5 & 37.2
        & 0.88
        & 1.16
    \\
     & 128
        & 48.1 & 20.5 & 28.6 & 30.4
        & \textbf{61.7} & \textbf{29.3} & \textbf{30.8} & \textbf{38.2}
        & 0.83
        & 1.26
    \\ \midrule
    512 & 8
        & - & - & - & -
        & - & - & - & -
        & - 
        & - 
    \\
     & 32
        & 46.7 & 19.7 & 28.1 & 29.6
        & 59.8 & 28.2 & 29.8 & 36.9
        & 0.77
        & 1.35
    \\
     & 64
        & 47.2 & 20.1 & 28.2 & 29.9
        & 61.1 & 29.3 & 30.7 & 38.0
        & 0.75
        & 1.40
    \\
     & 128
        & 47.2 & 20.0 & 28.2 & 29.9
        & 61.0 & 29.3 & 30.7 & 38.0
        & 0.71 
        & 1.51
    \\
    \bottomrule
\end{tabular}%
\caption{
  Varying the block size and the number of global tokens of a GlobalLocal encoder.
  Training steps per second and memory are computed based on arXiv, and normalized to the run with Block Size=128 and Global Tokens=32.
}
\label{tab:table_3_global_block_and_tokens}
\end{table*}

Next, we vary the block size and number of global tokens for the Global-Local encoder, with results shown in Table~\ref{tab:table_3_global_block_and_tokens}.\footnote{A number of experiments with very small block sizes or number global tokens ran into memory issues, owing to the way in which TPUs pad small dimensions of arrays to certain minimum lengths, leading to larger than expected memory consumption.}

Broadly, we find that increasing either block size or global tokens leads to improved performance, with a corresponding increase in memory consumption and computation time.
However, the effect size from going to larger block sizes is not large, and appears to saturate as we get to larger block sizes or number of global tokens.
As such, increasing either of these hyperparameters is preferable if resources allow, but may not be a high priority compared to other potential model improvements.
For the remainder of the ablation experiments, we stick to a block size of 64 and 32 global tokens for consistency.

\paragraph{Takeaways:} Larger block sizes and/or number of global tokens leads to improved performance, although the effect saturates.


\subsection{Position Encoding Schemes}

\begin{table*}[th]
\centering
\small
\resizebox{\textwidth}{!}{%
\begin{tabular}{l \rougecolumns \rougecolumns \rougecolumns \rougecolumns c}
    \toprule
    & \multicolumn{4}{c}{XSUM} 
    & \multicolumn{4}{c}{CNN/DM}
    & \multicolumn{4}{c}{arXiv} 
    & \multicolumn{4}{c}{GovReport}
    \\
    \cmidrule(lr){2-5}\cmidrule(lr){6-9}\cmidrule(lr){10-13}\cmidrule(lr){14-17}
    Position Encoding
    \rougecolumnnames
    \rougecolumnnames
    \rougecolumnnames
    \rougecolumnnames
    & Step/s
    \\ \midrule
    None
        & 34.3 & 12.5 & 26.8 & 22.6
        & 25.6 & 7.8 & 17.7 & 15.2
        & 36.1 & 9.8 & 22.0 & 19.8
        & 38.3 & 13.2 & 18.7 & 21.1
        & 0.96
    \\
    Sinusoidal
        & 39.8 & 16.9 & 31.8 & 27.8
        & \textbf{40.0} & \textbf{18.6} & \textbf{28.4} & \textbf{27.6}
        & 44.5 & 17.6 & 26.7 & 27.6
        & 40.0 & 18.8 & 22.3 & 25.6
        & 0.96
    \\
    T5
        & \textbf{40.1} & \textbf{17.1} & \textbf{32.0} & \textbf{28.0}
        & 39.8 & 18.8 & 28.6 & 27.8
        & \textbf{44.9} & \textbf{17.9} & \textbf{26.8} & \textbf{27.8}
        & \textbf{40.2} & \textbf{19.5} & \textbf{22.9} & \textbf{26.2}
        & 0.53
    \\
    RoPE
        & 39.8 & 16.9 & 31.8 & 27.8
        & 39.2 & 18.7 & 28.5 & 27.5
        & 43.5 & 17.2 & 26.5 & 27.1
        & 40.0 & 19.1 & 22.6 & 25.8
        & 0.85
    \\
    Absolute
        & 39.1 & 16.4 & 31.3 & 27.2
        & 39.7 & 18.7 & 28.5 & 27.7
        & 44.3 & 17.5 & 26.5 & 27.4
        & 38.6 & 17.5 & 21.1 & 24.2
        & \textbf{1.00}
    \\
    \bottomrule
\end{tabular}%
}
\caption{
  Comparison of position encodings schemes for a Transformer encoder-decoder.
  Training steps per second are computed based on arXiv summarization.
  Absolute position embeddings are replicated to longer input sequences, following \citet{beltagy2020longformer}.
  Training steps per second is computed based on arXiv, and normalized to the run with absolute position embeddings.
}
\label{tab:table_4_position_encoding}
\end{table*}

\begin{table*}[th]
\centering
\small
\resizebox{\textwidth}{!}{%
\begin{tabular}{l rr \rougecolumns \rougecolumns \rougecolumns \rougecolumns c}
    \toprule
    & & 
    & \multicolumn{4}{c}{XSUM} 
    & \multicolumn{4}{c}{CNN/DM}
    & \multicolumn{4}{c}{arXiv} 
    & \multicolumn{4}{c}{GovReport}
    \\
    \cmidrule(lr){4-7}\cmidrule(lr){8-11}\cmidrule(lr){12-15}\cmidrule(lr){16-19}
    Architecture & Enc & Dec
    \rougecolumnnames
    \rougecolumnnames
    \rougecolumnnames
    \rougecolumnnames
    \\ \midrule
    Local & 18 & 6
        & 37.4 & 15.0 & 29.7 & 25.5
        & 39.0 & 18.2 & 27.9 & 27.0
        & 46.0 & 19.4 & 27.6 & 29.1
        & 58.9 & 27.4 & 29.1 & 36.1
    \\
     & 12 & 12
        & 37.5 & 14.9 & 29.7 & 25.5
        & 38.5 & 18.0 & 27.6 & 26.7
        & 45.4 & 18.9 & 27.3 & 28.6
        & 59.2 & 27.6 & 29.3 & 36.3
    \\
     & 6 & 18
        & 37.7 & 15.1 & 29.9 & 25.7
        & 38.5 & 18.1 & 27.7 & 26.9
        & 46.3 & 19.3 & 27.6 & 29.1
        & 59.4 & 27.8 & 29.5 & 36.5
    \\ \midrule
    Global-Local & 18 & 6
        & \textbf{38.6} & \textbf{15.9} & \textbf{30.9} & \textbf{26.7}
        & 39.2 & 18.5 & 28.2 & 27.3
        & 47.3 & 20.1 & 28.3 & 30.0
        & 60.2 & 28.7 & 30.6 & 37.5
    \\
     & 12 & 12
        & 38.6 & 15.9 & 30.7 & 26.6
        & \textbf{40.0} & \textbf{18.6} & \textbf{28.3} & \textbf{27.6}
        & 47.5 & 20.1 & 28.3 & 30.0
        & \textbf{61.1} & \textbf{29.3} & \textbf{30.7} & \textbf{38.1}
    \\
     & 6 & 18
        & 37.7 & 15.1 & 29.9 & 25.7
        & 38.5 & 18.1 & 27.7 & 26.9
        & 46.4 & 19.5 & 27.9 & 29.3
        & 60.3 & 28.6 & 30.0 & 37.2
    \\ \midrule
    Global-Local & 18 & 12
        & 38.5 & 15.7 & 30.6 & 26.4
        & 38.7 & 18.4 & 28.1 & 27.1
        & 47.3 & 20.0 & 28.3 & 29.9
        & 60.2 & 29.2 & 31.0 & 37.9
    \\
     & 12 & 18
        & 38.6 & 15.8 & 30.5 & 26.5
        & 38.6 & 18.3 & 28.0 & 27.0
        & \textbf{47.5} & \textbf{20.3} & \textbf{28.5} & \textbf{30.2}
        & 60.9 & 29.0 & 30.4 & 37.7
    \\

    \bottomrule
\end{tabular}%
}
\caption{
    Varying the distribution of encoder/decoder layers)
}
\label{tab:table_5_scaling}
\end{table*}
\label{experiments_position_encoding}

New position encoding schemes encoding schemes such as RoPE \citep{su2021rope} and ALiBi \citep{press2022alibi} have garnered recent attention, showing improved performance on downstream evaluations.
As input sequence lengths have gotten much longer, and in particular longer than the dimensions of hidden representations, previous choices of position encoding may no longer be optimal.
Moreover, relative position encodings such as RoPE, T5 and ALiBi may be better suited for adapting models to different input lengths between pretraining and fine-tuning.
Hence, this is a good opportunity to revisit the choice of positioning encoding schemes in encoder models.

Because of the more complex interaction between local attention blocks and relative position encoding implementations, we conduct a preliminary investigation with a full-attention Transformer.
We pretrain with an input length of 512, and fine-tune with an input length of 2048 for the long sequence tasks -- this experiment also tests the propensity for position encodings to be adapted to longer sequences downstream.
In addition to the sinusoidal position encoding used in PEGASUS and \citet{vaswani2017transformer}, we also consider the bucket-based relative position encoding scheme of T5, RoPE, absolute position embeddings, and no position encoding as a baseline.
For absolute position embeddings, we follow the recipe of \citet{beltagy2020longformer} and duplicate the learned position embeddings to handle longer sequences before fine-tuning.
The chosen position encoding scheme is applied to all parts of the model, including both the encoder and the decoder.
We do not experiment with ALiBi, as we found no natural way to adapt ALiBi to cross-attention.

Our results are shown in Table~\ref{tab:table_4_position_encoding}.
We find that although T5 performs the best, it is also almost twice as slow as the other position encoding schemes, which is consistent with the findings of \citet{press2022alibi}.
Sinusoidal position encodings and RoPE perform only slightly worse than T5 with much better efficiency, making them more desirable choices.
Given the much simpler implementation of sinusoidal position encodings, we opt to stick with them for the remainder of the experiments.

\paragraph{Takeaways:} Sinusoidal position encodings still remain a good choice for long input Transformers.



\subsection{Scaling Encoder and Decoder Layers}


Scaling laws \citep{kaplan2020scaling,ghorbani2021scaling,zhang2022examining} that describe the empirical relationship between model sizes and performance have proven surprisingly consistent and gotten significant attention in recent years.
We present in this section a small set of scaling experiments, exploring the distribution of layers between encoder and decoder. 

Our results are shown in Table \ref{tab:table_5_scaling}.
In the top half, we fix the total number of layers to 24, and consider both encoder-heavy and decoder-heavy distributions, for both Local and Global-Local models. 
We observe that impact of distribution of encoder and decoder layers on performance is relatively small.
For Local models, we see a slight boost from decoder-heavy models.
For Global-Local models, we observe that a balanced encoder-decoder outperforms encoder- and decoder-heavy models, both of which perform about comparably.

\begin{table*}[th]
\centering
\small
\begin{tabular}{ll \rougecolumns \rougecolumns}
    \toprule
    &
    & \multicolumn{4}{c}{arXiv} 
    & \multicolumn{4}{c}{GovReport}
    \\
    \cmidrule(lr){3-6}\cmidrule(lr){7-10}
    Pretraining $\rightarrow$ Fine-tuning
    & Block Size
    \rougecolumnnames
    \rougecolumnnames
    \\ \midrule
    Transformer $\rightarrow$ Local & 4
        & 46.2 & 19.6 & 27.9 & 29.3
        & 60.0 & 28.3 & 29.8 & 37.0
    \\
     & 16
        & 46.4 & 19.6 & 27.9 & 29.4
        & 59.6 & 28.2 & 29.9 & 36.9
    \\
     & 64
        & 46.5 & 19.5 & 27.8 & 29.3
        & 59.5 & 28.0 & 29.6 & 36.7
    \\
     & 256
        & 46.8 & 19.7 & 28.0 & 29.6
        & 59.8 & 28.0 & 29.8 & 36.8
    \\
    Local $\rightarrow$ Local & 4
        & 45.0 & 18.2 & 26.6 & 27.9
        & 59.1 & 27.1 & 28.8 & 35.9
    \\
     & 16
        & 45.9 & 19.1 & 27.5 & 28.9
        & 59.0 & 27.5 & 29.3 & 36.2
    \\
     & 64
        & 46.5 & 19.5 & 27.8 & 29.3
        & 59.7 & 28.1 & 29.8 & 36.8
    \\
     & 256
        & 47.1 & 19.9 & 28.1 & 29.8
        & 59.7 & 28.5 & 30.3 & 37.2
    \\
    Transformer $\rightarrow$ Global-Local & 4
        & 44.6 & 18.0 & 26.6 & 27.7
        & 59.5 & 27.0 & 28.6 & 35.8
    \\
     & 16
        & 46.0 & 19.2 & 27.5 & 29.0
        & 60.3 & 28.2 & 29.8 & 37.0
    \\
     & 64
        & 47.0 & 20.0 & 28.2 & 29.8
        & 60.8 & 28.7 & 30.1 & 37.4
    \\
     & 256
        & 47.6 & 20.3 & 28.4 & 30.2
        & 60.8 & 28.7 & 30.0 & 37.4
    \\
    Global-Local $\rightarrow$ Global-Local & 4
        & 46.1 & 18.8 & 27.2 & 28.7
        & 60.1 & 27.6 & 28.9 & 36.3
    \\
     & 16
        & 47.1 & 20.0 & 28.3 & 29.9
        & 59.7 & 27.8 & 29.2 & 36.5
    \\
     & 64
        & \textbf{47.7} & \textbf{20.3} & \textbf{28.5} & \textbf{30.2}
        & 61.0 & 29.3 & 30.8 & 38.0
    \\
     & 256
        & 47.3 & 20.2 & 28.3 & 30.0
        & \textbf{61.6} & \textbf{29.4} & \textbf{30.7} & \textbf{38.2}
    \\
    \bottomrule
\end{tabular}%
\caption{
    Comparison of adapting models architectures between pretraining and fine-tuning.
}
\label{tab:table_6_pretrain_vs_finetune}
\end{table*}


We also consider cases where we further increase the size of either the encoder or decoder to 18 layers, shown in the second half of Table~\ref{tab:table_5_scaling}.
We observe no improvement in performance over the 12/12-layer encoder-decoder, and suspect that other hyperparameters (e.g. hidden size) might be the bottleneck rather than the number of layers.

We highlight here that because of the asymmetry of the input and output lengths, there are different computational trade-offs to different balances of encoder and decoder layers. 
Encoder-heavy models require more memory because of the long input sequences, whereas decoder-heavy models are relative slower at inference because of the autoregressive nature of decoding.
Given the relatively small difference in the margin of performance, memory or computational constraints may outweigh the performance differences in practical scenarios.

\paragraph{Takeaways:} A balanced Global-Local model outperforms other variants, but the difference in performance may be outweighed by other resource considerations.


\subsection{Pretraining vs Fine-tuning Architectures}

Previous works using efficient Transformer encoders have generally taken the model weights of a full-attention Transformer pretrained on a shorter sequence, and adapted them to the efficient architecture either directly during fine-tuning \citep{zaheer2020bigbird}, or with an intermediate stage of additional pretraining \citep{beltagy2020longformer}.
In this section, we investigate if such an approach is optimal, or if the model would benefit from being pretrained with the efficient encoder from the beginning.
Note that we are still performing pretraining on a short sequence (512 tokens), even with an efficient encoder.

We consider both pretraining with a Transformer and pretraining with the efficient architecture for both Local and Global-Local models.
We also vary the block size, as the main difference between a Transformer and Local Transformer is the block size (aside from staggering, a Local model with block size 512 is equivalent to a dense Transformer), and hence the difference in block size also corresponds to the extent to which the model needs to adapt between architectures.
When adapting from a pretrained Transformer encoder to a Global-Local architecture, because the Global-Local model relies on newly introduced global token embeddings, we initialize them by randomly sampling tokens from the vocabulary embeddings.

Our results are shown in Table~\ref{tab:table_6_pretrain_vs_finetune}.
For Local models, we find that pretraining with local attention using small block sizes tends to hurt performance, but at moderate block sizes (e.g. 64) there is little difference between the two approaches.
In contrast, we find that for Global-Local, pretraining with the efficient architecture tends to perform better.
We hypothesize that this difference arises because of the presence of the learned global embedding tokens, which are randomly initialized when adapting from a pretrained Transformer and hence may benefit from pretraining and being jointly trained with the local attention.

\paragraph{Takeaways:} For moderate block sizes, either pretraining or adapting to a Local encoder performs about equally well, but pretraining with a Global-Local encoder performs slightly better.


\subsection{Pretraining Schemes}

\begin{table*}[th]
\centering
\small
\resizebox{\textwidth}{!}{%
\begin{tabular}{ll \rougecolumns \rougecolumns \rougecolumns \rougecolumns}
    \toprule
    &
    & \multicolumn{4}{c}{XSUM} 
    & \multicolumn{4}{c}{CNN/DM}
    & \multicolumn{4}{c}{arXiv} 
    & \multicolumn{4}{c}{GovReport}
    \\
    \cmidrule(lr){3-6}\cmidrule(lr){7-10}\cmidrule(lr){11-14}\cmidrule(lr){15-18}
    Pretraining Scheme
    & Encoder
    \rougecolumnnames
    \rougecolumnnames
    \rougecolumnnames
    \rougecolumnnames
    \\ \midrule
    Short (50\%) & Local
        & 38.4 & 15.8 & 30.6 & 26.5
        & 39.2 & 18.1 & 27.9 & 27.1
        & 46.8 & 19.7 & 28.0 & 29.6
        & 60.1 & 28.3 & 29.8 & 37.0
    \\
     & Global-Local
        & 39.4 & 16.5 & 31.5 & 27.4
        & 39.1 & 18.6 & 28.3 & 27.4
        & 47.7 & 20.4 & 28.6 & 30.3
        & 61.9 & 29.6 & 30.8 & 38.4
    \\
    Short (100\%) & Local
        & 39.2 & 16.3 & 31.3 & 27.1
        & 39.2 & 18.6 & 28.3 & 27.4
        & 46.9 & 19.7 & 28.0 & 29.6
        & 60.1 & 28.3 & 29.8 & 37.0
    \\
     & Global-Local
        & \textbf{39.9} & \textbf{17.0} & \textbf{31.9} & \textbf{27.9}
        & 39.8 & 18.6 & 28.3 & 27.6
        & 48.1 & 20.5 & 28.7 & 30.5
        & 61.9 & 29.6 & 30.8 & 38.4
    \\
    Short (75\%) $\rightarrow$ Long (25\%) & Local
        & 38.8 & 15.9 & 30.7 & 26.7
        & 39.1 & 18.2 & 28.0 & 27.1
        & 47.5 & 20.1 & 28.2 & 30.0
        & 60.6 & 28.9 & 30.6 & 37.7
    \\
     & Global-Local
        & 39.6 & 16.8 & 31.7 & 27.6
        & \textbf{39.8} & \textbf{18.8} & \textbf{28.5} & \textbf{27.7}
        & 48.4 & 20.7 & 28.8 & 30.7
        & 61.8 & 29.8 & 31.1 & 38.5
    \\
    Short (50\%) $\rightarrow$ Long (50\%) & Local
        & 38.4 & 15.7 & 30.5 & 26.4
        & 39.4 & 18.1 & 27.9 & 27.1
        & 47.7 & 20.2 & 28.3 & 30.1
        & 60.9 & 29.1 & 30.7 & 37.9
    \\
     & Global-Local
        & 39.3 & 16.4 & 31.4 & 27.3
        & 39.4 & 18.3 & 28.1 & 27.3
        & \textbf{48.4} & \textbf{20.9} & \textbf{29.1} & \textbf{30.9}
        & \textbf{61.7} & \textbf{30.0} & \textbf{31.2} & \textbf{38.7}
    \\
    Long (100\%) & Local
        & 36.0 & 14.0 & 28.6 & 24.3
        & 38.4 & 17.7 & 27.4 & 26.5
        & 46.7 & 19.5 & 27.7 & 29.3
        & 59.8 & 28.0 & 29.5 & 36.7
    \\
     & Global-Local
        & 36.4 & 14.3 & 28.9 & 24.7
        & 38.5 & 17.8 & 27.5 & 26.6
        & 47.3 & 19.9 & 28.1 & 29.8
        & 61.1 & 29.1 & 30.7 & 37.9
    \\
    \bottomrule
\end{tabular}%
}
\caption{
  Comparison of different pretraining formats, given a input token budget of 131B tokens, which corresponds to 1M steps with 512 input tokens.
  Short pretraining uses 512 input tokens, whereas long pretraining uses 4096 input tokens.
}
\label{tab:table_7_pretraining_setups}
\end{table*}

Up to this point, we have only considered pretraining with short sequences.
We might expect that pretraining with longer sequences ought to improve performance of our model on downstream long input summarization.
However, pretraining only on long sequences is computationally expensive and requires a large collection of long input documents, which are relatively rarer.
Moreover, long documents may contain different information from short documents, hence limiting training to only long inputs maybe reduce the diversity of training data.
Different long context Transformers have taken different approaches to pretraining on long inputs.
For instance, Longformer \citep{beltagy2020longformer} performed several additional stages of increasingly longer-sequence pretraining to adapt the initial RoBERTa to long sequence inputs.
On the other hand, LongT5 \citep{guo2021longt5} is pretrained exclusively with long input sequences.
Others \citep{zaheer2020bigbird, ivgi2022sled} perform no long input pretraining at all.
In this section, we investigate how the balance of short and long pretraining impact downstream performance, and try to find the best trade-off between pretraining cost and downstream performance.

We consider two setups for pretraining: \textit{short-input pretraining}, with 512 input tokens and 256 output tokens, and \textit{long-input pretraining}, with 4096 input tokens and 256 output tokens.
We describe the corresponding differences in data preprocessing in Section~\ref{sec:exp_pretraining}.
We choose to fix the number of input tokens seen during training as the constraint, and vary configurations subject to this constraint.
This constraint roughly proxies for the amount of compute consumed as well as corresponds to the number of input tokens seen during pretraining, in contrast to fixing the number of steps, where long-input pretraining would consume far more compute for the same number of steps.

In contrast to the above experiments where we generally performed short pretraining for 500k steps, we set our total input token budget at 131 billion tokens, which correponds to 1 million steps with 512 input tokens.
This larger budget ensures that when we do only long-input pretraining, the model is still pretrained for a reasonable number of steps.
Given this budget, we consider four configurations:

\begin{itemize}
    \item Short-input pretraining for 100\% of tokens (1M steps)
    \item Short-input for 75\% of tokens (98.3B, 750k steps), then long-input for 25\% of tokens (32.8B, 31.25k steps)
    \item Short-input for 50\% of tokens (62.5B, 500k steps), then long-input for 50\% of tokens (62.5B, 62.5k steps)
    \item Long-input pretraining for 100\% of tokens (125k steps)
\end{itemize}

\begin{table*}[th]
\centering
\small
\resizebox{\textwidth}{!}{%
\begin{tabular}{l \rougecolumns \rougecolumns \rougecolumns \rougecolumns cc}
    \toprule
    & \multicolumn{4}{c}{XSUM} 
    & \multicolumn{4}{c}{CNN/DM}
    & \multicolumn{4}{c}{arXiv} 
    & \multicolumn{4}{c}{GovReport}
    \\
    \cmidrule(lr){2-5}\cmidrule(lr){6-9}\cmidrule(lr){10-13}\cmidrule(lr){14-17}
    Cross-Attention
    \rougecolumnnames
    \rougecolumnnames
    \rougecolumnnames
    \rougecolumnnames
    & Step/s & Mem
    \\ \midrule
    Full
        & \textbf{38.8} & \textbf{16.0} & \textbf{31.0} & \textbf{26.8}
        & 39.5 & 18.6 & 28.4 & 27.5
        & 47.7 & 20.4 & 28.6 & 30.3
        & \textbf{61.3} & \textbf{29.4} & \textbf{30.8} & \textbf{38.1}
        & 1.00
        & 1.00
    \\
    Cross[0,2,4,6,8,10]
        & 38.3 & 15.6 & 30.5 & 26.3
        & \textbf{39.8} & \textbf{18.8} & \textbf{28.5} & \textbf{27.7}
        & \textbf{48.1} & \textbf{20.4} & \textbf{28.6} & \textbf{30.4}
        & 61.0 & 29.0 & 30.7 & 37.9
        & 1.10
        & 0.90
    \\
    Cross[0,3,6,9,11]
        & 38.0 & 15.3 & 30.2 & 26.0
        & 38.8 & 18.4 & 28.1 & 27.2
        & 46.9 & 19.9 & 28.2 & 29.7
        & 60.1 & 28.6 & 30.2 & 37.3
        & 1.15
        & 0.88
    \\
    Cross[0,4,8,11]
        & 37.8 & 15.3 & 30.1 & 25.9
        & 38.5 & 18.1 & 27.9 & 26.9
        & 47.6 & 20.2 & 28.4 & 30.1
        & 60.9 & 28.9 & 30.3 & 37.6
        & 1.15
        & 0.86
    \\
    Cross[0,6,11]
        & 37.4 & 14.8 & 29.7 & 25.4
        & 38.8 & 18.1 & 27.9 & 27.0
        & 46.9 & 19.7 & 28.1 & 29.6
        & 60.3 & 28.5 & 30.2 & 37.3
        & 1.18
        & 0.87
    \\
    Cross[0,6]
        & 37.5 & 14.9 & 29.7 & 25.5
        & 38.3 & 18.0 & 27.8 & 26.8
        & 47.1 & 19.8 & 28.1 & 29.7
        & 60.4 & 28.1 & 29.7 & 36.9
        & \textbf{1.21}
        & \textbf{0.85}
    \\

    \bottomrule
\end{tabular}%
}
\caption{
  Comparison of models with cross-attention only in a subset of the 12 decoder layers.
  Training steps per second and memory are computed based on arXiv, and normalized to the Cross[0,6] run.
}
\label{tab:table_8_partial_cross}
\end{table*}

We compare the performance of the different pretraining scehemes in Table~\ref{tab:table_7_pretraining_setups}.
We also include the short-input pretraining for 500k steps for comparison.
First, comparing short-input pretraining for 500k and 1M steps, we find that more pretraining still improves performance, indicating that our base models may still be undertrained at 500k steps.
Secondly, we observe that long-input pretraining performs consistently worse than the other variants, which we attribute to the fewer number of training steps taken, again highlighting the issue of potential under-training.
Focusing our analysis on the middle three configurations, on the long tasks, we find that all three non-long-only variants atttain similar scores, with more long-input pretraining having slightly better performance, particularly on the ROUGE-2 and ROUGE-L scores.
While the small absolute differences in scores make it hard to draw strong conclusions, we lean towards the conclusion that adding a short phase of long input pretraining can be beneficial can improve performance on long input summarization tasks.\footnote{One major difference from Longformer is that Longformer uses absolute position embeddings, hence it is potentially more important the model to have some pretraining with longer sequences to adapt the replicated position embeddings to capture different position information. In contrast, because our models use sinusoidal position encodings which can naturally extrapolate to longer input lengths, we find that fine-tuning has been sufficient to adapt the model to reasonable performance.}

\begin{table*}[th]
\centering
\small
\begin{tabular}{ll \rougecolumns \rougecolumns}
    \toprule
    &
    & \multicolumn{4}{c}{arXiv} 
    & \multicolumn{4}{c}{GovReport}
    \\
    \cmidrule(lr){3-6}\cmidrule(lr){7-10}
    Cross-Attention
    & Model
    \rougecolumnnames
    \rougecolumnnames
    \\ \midrule
    Pretrained & Full
        & 47.7 & 20.4 & 28.6 & 30.3
        & \textbf{61.3} & \textbf{29.4} & \textbf{30.8} & \textbf{38.1}
    \\
    & Cross[0,2,4,6,8,10]
        & \textbf{48.1} & \textbf{20.4} & \textbf{28.6} & \textbf{30.4}
        & 61.0 & 29.0 & 30.7 & 37.9
    \\
     & Cross[0,6]
        & 47.1 & 19.8 & 28.1 & 29.7
        & 60.4 & 28.1 & 29.7 & 36.9
    \\
    Converted & Cross[0,2,4,6,8,10]
        & 46.4 & 19.7 & 28.1 & 29.5
        & 60.2 & 28.8 & 30.3 & 37.4
    \\
     & Cross[0,6]
        & 46.2 & 19.7 & 28.1 & 29.5
        & 60.2 & 28.1 & 29.8 & 36.9
    \\
    \bottomrule
\end{tabular}%
\caption{
  Comparison of models pretrained with cross-attention for a subset of layers, and adapting a pretrained model by dropping cross-attention layers only during fine-tuning
}
\label{tab:table_9_partial_converted}
\end{table*}

\paragraph{Takeaways:} Given a fixed compute budget, allocating some portion of training to long-input training can improve performance, although the precise optimal allocation is difficult to determine. Exclusively long pretraining results in worse performance.


\subsection{Partial Cross Attention}

Given the use of an efficient attention architecture, which has memory consumption scale linearly rather than quadratically in input sequence length, another major memory bottleneck is the encoder-decoder cross-attention.
Because each decoder layer attends separately to the long encoder representations, and the attention is dense, this is a large contiguous chunk of memory that we could seek to reduce.

Perceiver AR \citep{hawthorne2022perceiverar} demonstrated strong performance by using only a single cross-attention at the bottom layer of an autoregressive language model.
Based on these results, we investigate the impact of only having cross-attention on a subset of decoder layers. 
In Table~\ref{tab:table_8_partial_cross}, we show the results of pretraining and fine-tuning Global-Local models with cross-attention only on specific layers on a variety of configurations.
We find that reducing the number of cross-attention layers leads to a drop in performance, but the impact on performance is smaller than expected.
For instance, with only cross-attention on the first and sixth layer, the Global-Local model still outperforms a Local model.
The reduction of cross-attention layers also leads to a corresponding improvement in training step and reduction in memory consumption.

Given the small drop in performance from using fewer decoder layers with cross-attention, we consider the viability of dropping cross-attention layers after pretraining.
In other words, we take a Global-Local model pretrained with full cross-attention, drop the cross-attention for a subset of layers, and fine-tune directly.
Our results are shown in Table~\ref{tab:table_9_partial_converted}.
We find that dropping the cross-attention after pretraining again only leads to a small (additional) dip in performance.
This indicates that dropping cross-attention may be a viable strategy for further reducing memory requirements for an existing pretrained model with a small performance trade-off, and pretraining a separate model from scratch is not necessary.

\paragraph{Takeaways:} Dropping cross-attention for a fraction of decoder layers can reduce memory consumption at the cost of slight performance regression. Cross-attention can be dropped after pretraining, with an associated performance trade-off.





\section{\largemodel{}}

Based on our findings above, we settle on the following recipe for adapting the PEGASUS models \citep{zhang2019pegasus} to long sequence summarization. 

\begin{itemize}
    \item We use a Global-Local architecture with block staggering, a large number of global tokens, and large block sizes during pretraining.
    \item We conduct an additional stage of long input pretraining on 4096 token inputs for 300,000 steps.
    \item We extend input sequences up to 16384 input tokens in fine-tuning, depending on the task.
\end{itemize}

We experiment with two model sizes \textbf{\largemodel{}} (PEGASUS e\textbf{X}tended), based on \pegasuslarge; and
\smallmodel{}, based on a newly trained \pegasusbase{} model which we call \pegasusbaseplus{}.
In a similar finding as \citet{chinchilla}, we found that \pegasusbase{}  benefits from training on significantly more tokens, which we set to the same as \pegasuslarge{}.

We initialize the weights of \largemodel{} and \smallmodel{} on the pretrained weights of \pegasuslarge{} and \pegasusbaseplus{} respectively.
Only two new sets of parameters introduced: the global token embeddings, and a separate LayerNorm for the global input representations in each Transformer layer.
This is approximately 1M more parameters for \smallmodel{} and 2M more for \largemodel{}.
We initialize the global token embeddings by randomly sampling tokens from the input token embedding, and we initialize the LayerNorm weights with the regular input LayerNorm weights.

The task- and model-specific hyperparameters for fine-tuning can be found in Appendix~\ref{tab:table_a2_finetuning_hyperparameters}. For this section, we report ROUGE-Lsum\footnote{\url{https://github.com/google-research/google-research/blob/master/rouge/README.md\#two-flavors-of-rouge-l}} rather than ROUGE-L for consistency with the metrics reported in other papers and leaderboards.

\begin{table}[th]
\centering
\small
\begin{tabular}{l cc}
    \toprule
    & \smallmodel
    & \largemodel
    \\ \midrule
    \# Parameters & 272M & 568M
    \\
    \# Global Tokens & 128 & 128
    \\
    Block Size & 512 & 512
    \\
    Batch Size & 512 & 1024
    \\
    \multirow{2}{*}{{\shortstack[l]{ Additional \\ Pretraining}}}
    & \multirow{2}{*}{{\shortstack[c]{ 300K steps}}}
    & \multirow{2}{*}{{\shortstack[c]{ 300K steps}}}
    \\
    \
    \\
    \bottomrule
\end{tabular}%
\caption{
  Hyperparameters of Pegasus-X Models
}
\label{tab:table_10_pegasusx_hyperparams}
\end{table}

\subsection{Results on Summarization tasks}

\begin{table*}[th]
\centering
\small
\resizebox{\textwidth}{!}{%
\begin{tabular}{l c \rougecolumns \rougecolumns \rougecolumns}
    \toprule
    &
    & \multicolumn{4}{c}{arXiv} 
    & \multicolumn{4}{c}{Big Patent}
    & \multicolumn{4}{c}{PubMed}
    \\
    \cmidrule(lr){3-6}
    \cmidrule(lr){7-10}
    \cmidrule(lr){11-14}
    Model
    & \#Params
    \rougecolumnnameslsum
    \rougecolumnnameslsum
    \rougecolumnnameslsum
    \\ \midrule
    \pegasusbase & 271M
        & 34.8 & 10.2 & 22.5* & 20.0*
        & 43.5 & 20.4 & 31.8* & 30.5*
        & 40.0 & 15.2 & 25.2* & 24.8*
    \\
    \pegasusbaseplus & 271M
        & 42.2 & 15.8 & 37.3 & 29.2
        & 51.2 & 32.6 & 41.0 & 40.9
        & 44.1 & 18.3 & 40.1 & 31.9
    \\
    \pegasusbaseplusgl & 272M
        & 47.6 & 20.2 & 42.4 & 34.4
        & 58.1 & 39.5 & 47.2 & 47.7
        & 47.3 & 21.4 & 43.0 & 35.2
    \\
    \smallmodel & 272M
        & 49.4 & 21.6 & 44.0 & 36.1
        & 61.3 & 42.6 & 50.1 & 50.8
        & 49.6 & 23.6 & 45.2 & 37.5
    \\
    \midrule
    \pegasuslarge & 567M
        & 44.7 & 17.2 & 25.7* & 27.0*
        & 53.4 & 32.9 & 42.1* & 42.0*
        & 45.1 & 19.6 & 27.4* & 28.9*
    \\
    \largemodel & 568M
        & 50.0 & 21.8 & 44.6 & 36.5
        & 64.8 & 47.5 & 54.3 & 55.1
        & \textbf{51.0} & \textbf{24.7} & \textbf{46.6} & \textbf{38.9}
    \\
    \midrule
    \midrule
    Longformer Encoder-Decoder & 464M 
        & 46.6 & 19.6 & 41.8 & 33.7
        & --.- & --.- & --.- & --.-
        & --.- & --.- & --.- & --.-
    \\
    Top-Down (AvgP) & 464M 
        & 48.7 & 20.7 & 43.9 & 35.4
        & --.- & --.- & --.- & --.-
        & 48.3 & 21.4 & 44.2 & 35.7
    \\
    Top-Down (AdaP) & 464M 
        & \textbf{51.0} & \textbf{21.9} & \textbf{45.6} & \textbf{37.1}
        & --.- & --.- & --.- & --.-
        & 51.1 & 23.3 & 46.5 & 38.1
    \\
    Big Bird-Pegasus & 567M
        & 46.6 & 19.0 & 41.8 & 33.3
        & 60.6 & 42.5 & 50.1 & 50.5
        & 46.3 & 20.7 & 42.3 & 34.4
    \\
    LongT5\textsubscript{Large} & 770M
        & 48.3 & 21.6 & 44.1 & 35.8
        & 70.4 & 56.8 & 62.7 & 63.1
        & 50.0 & 24.7 & 46.5 & 38.6
    \\
    LongT5\textsubscript{XL} & 3B
        & 48.4 & 21.9 & 44.3 & 36.1
        & \textbf{76.9} & \textbf{66.1} & \textbf{70.8} & \textbf{71.1}
        & 50.2 & 24.8 & 46.7 & 38.7
    \\
    \bottomrule
\end{tabular}%
}
\caption{
  Comparison on long summarization tasks (Test sets). Results for other models are taken from their respective papers.
  *: PEGASUS \citep{zhang2019pegasus} only reports ROUGE-L and not ROUGE-LSum.
}
\label{tab:table_11_pegasusx_long}
\end{table*}

\begin{table*}[th]
\centering
\small
\begin{tabular}{l \rougecolumns \rougecolumns \rougecolumns \rougecolumns}
    \toprule
    & \multicolumn{4}{c}{CNN/DailyMail}
    & \multicolumn{4}{c}{XSum} 
    \\
    \cmidrule(lr){2-5}
    \cmidrule(lr){6-9}
    Model
    \rougecolumnnameslsum
    \rougecolumnnameslsum
    \\ \midrule
    \pegasusbase
        & 41.8 & 18.8 & 38.9 & 38.9
        & 39.8 & 16.6 & 31.7 & 27.6
    \\
    \pegasusbaseplus
        & 42.5 & 20.1 & 39.6 & 32.4
        & 43.8 & 21.2 & 36.0 & 32.2
    \\
    \smallmodel
        & 42.5 & 20.1 & 39.6 & 32.4
        & 42.9 & 20.1 & 35.0 & 31.2
    \\
    \midrule
    \pegasuslarge
        & \textbf{44.2} & \textbf{21.5} & \textbf{41.1} & \textbf{33.9}
        & \textbf{47.2} & \textbf{24.6} & \textbf{39.2} & \textbf{35.7}
    \\
    \largemodel
        & 43.4 & 21.2 & 40.6 & 33.5
        & 45.8 & 22.8 & 37.6 & 34.0
    \\
    \bottomrule
\end{tabular}%
\caption{
  Comparison on short summarization tasks (Test sets)
}
\label{tab:table_12_pegasusx_short}
\end{table*}

\paragraph{Long summarization tasks}

In Table~\ref{tab:table_11_pegasusx_long}, we compare the performance of PEGASUS models to those of \largemodel{} on three long-input summarization tasks: arXiv, Big Patent and PubMed.
In all three tasks, we see significant improvements in performance of \smallmodel{} over \pegasusbaseplus{}, and \largemodel{} over \pegasuslarge.
To isolate the impact of additional long input pretraining compared to only switching the architecture to accomodate long input sequences, we also include evaluation on the PEGASUS models using the Global-Local architecture with no further pretraining, which we list in the table as \pegasusbaseplus{} + Global-Local.

We also compare to reported results of \pegasuslarge{} using the Big Bird architecture \citep{zaheer2020bigbird}, Longformer encoder-ecoder \citep[LED;][]{beltagy2020longformer}, the Top-Down Transformer \citep{pang2022topdown} in both Average-Pool (AvgP) and Adaptive-Pool (AdaP) variants, the Large and XL sizes of LongT5, and the SLED \citep{ivgi2022sled}. LED, Top-Down and SLED are all initialized with BART\textsubscript{Large} weights with no additional pretraining on long input sequences, although AdaP has a multi-step fine-tuning setup (see below).

We note that the Big Bird-PEGASUS uses only 3072 tokens context, which is likely due to the larger memory consumption of Big Bird.
We find that \largemodel{} outperforms Big Bird-PEGASUS on all tasks, and Top-Down-AvgP on both compared tasks.
Top-Down-AdaP still outperforms \largemodel{}, we highlight that Top-Down-AdaP uses a much more complex, multi-step fine-tuning setup, involving using an importance tagger on reference summaries to construct weights for pooling tokens within segments.
In contrast, \largemodel{} is fine-tuned with the standard fine-tuning pipeline.
Even so, \largemodel{} still outperforms Top-Down with adaptive pooling on PubMed.
\largemodel{} also outperforms LongT5 on both arXiv and PubMed summarization, despite both compared LongT5 models having more parameters.
However, we find that LongT5 performs much better on BigPatent, which is a largely extractive summarization task. We hypothesize that a much larger much may be better at extraction over very long encoded sequences.

\paragraph{Short summarization tasks}

We show in Table~\ref{tab:table_12_pegasusx_short} the performance of PEGASUS and \largemodel{} models on shorter summarization tasks.
We observe that there is a slight regression in performance of both \largemodel{} models compared to their PEGASUS equivalents. 
We hypothesize that the long input pretraining might negatively impact the performance on shorter input tasks because of the difference data filtering for long documents, resulting in a potentially less diverse training data distribution.

\begin{table*}[th]
\centering
\small
\begin{tabular}{l c \rougecolumns \rougecolumns \rougecolumns}
    \toprule
    & 
    & \multicolumn{4}{c}{GovReport} 
    & \multicolumn{4}{c}{SummScreen/FD}
    & \multicolumn{4}{c}{QMSum}
    \\
    \cmidrule(lr){3-6}
    \cmidrule(lr){7-10}
    \cmidrule(lr){11-14}
    Model
    & \#Params
    \rougecolumnnames
    \rougecolumnnames
    \rougecolumnnames
    \\ \midrule
    \smallmodel & 272M
        & 59.3 & 29.3 & 30.9 & 37.7
        & 35.0 & 8.9 & 20.4 & 18.5
        & 32.9 & 9.8 & 21.4 & 19.0
    \\
    \largemodel & 568M
        & \textbf{60.3} & \textbf{30.0} & \textbf{31.5} & \textbf{38.5}
        & 35.7 & 9.1 & 20.6 & 18.8
        & 33.2 & 9.6 & 21.6 & 19.0
    \\ \midrule
    BART\textsubscript{Large}-SLED & 406M
        & 58.0 & 26.9 & 27.6 & 35.1
        & 33.8 & 8.0 & 18.5 & 17.1
        & 32.1 & 10.2 & 21.0 & 19.0
    \\
    Top-Down-AvgP & 464M 
        & --.- & --.- & --.- & --.-
        & 35.8 & 8.9 & 30.6* & 21.4*
        & --.- & --.- & --.- & --.-
    \\
    Top-Down-AdaP & 464M
        & --.- & --.- & --.- & --.-
        & \textbf{36.8} & \textbf{9.2} & \textbf{31.1*} & \textbf{21.9*}
        & --.- & --.- & --.- & --.-
    \\
    LongT5\textsubscript{Large} & 770M
        & 54.2 & 27.8 & 29.8 & 35.5
        & 35.6 & 9.2 & 21.2 & 19.1
        & \textbf{35.1} & \textbf{12.0} & \textbf{23.3} & \textbf{21.4}
    \\
    LongT5\textsubscript{XL} & 3B
        & 54.7 & 28.2 & 30.2 & 36.0
        & 35.8 & 9.6 & 21.1 & 19.4
        & 34.9 & 11.8 & 23.5 & 21.3
    \\
    UL2 & 20B
        & 53.6 & 26.1 & 28.8 & 34.3
        & 32.9 & 7.8 & 19.4 & 17.1
        & 31.1 & 8.5 & 20.4 & 17.5
    \\
    \bottomrule
\end{tabular}%
\caption{
  Comparison on SCROLLS benchmark (Summarization tasks, Test sets). Results for SLED, LongT5 and UL2 models are taken from the SCROLLS benchmark leaderboard.
  *: Top-Down \citep{pang2022topdown} reports much higher scores for ROUGE-L on SummScreen/FD than any other model, and may have been computed with a variant of ROUGE-L that involves splitting on sentences rather than newlines.
}
\label{tab:table_13_scrolls_sum}
\end{table*}

\subsection{Results on SCROLLS Summarization Tasks}


We report the performance of the \largemodel{} models on the summarization tasks in the recently introduced SCROLLS benchmark in Table~\ref{tab:table_13_scrolls_sum}. This includes GovReport \citep{huang2021govreport}, the ForeverDreaming subset of SummScreen \citep{chen2022summscreen}, and QMSum \citep{zhong2021qmsum}.

We observe that \largemodel{} outperforms all other models on GovReport, setting the state of the art on the dataset. 
\largemodel{} performs comparably to both LongT5\textsubscript{Large} and Top-Down-AvgP on SummScreen/FD, although it underperforms both LongT5 models on QMSum. 
Moreover, we find that \smallmodel{} also performs competitively, outperforming both LongT5 models on GovReport, and only a small margin behind \largemodel{} on all three tasks.
\smallmodel{} also outperforms BART\textsubscript{Large}-SLED, a larger model with a similar 16K token context length.
A major difference between \largemodel{} and BART\textsubscript{Large}-SLED, besides being based on PEGASUS and BART respectively, is that BART\textsubscript{Large}-SLED does not have additional pretraining on long documents.
We also note that UL2 only uses a context length of 2K tokens.



\section{Related Work}
\paragraph{Long Document Summarization} 
Several new long input summarization datasets and benchmarks have been recently introduced, providing better measures of long input summarization capability as well as prompting new interest in this research direction.
The BookSum dataset \citep{kryscinski2021booksum} consists of paragraph, chapter, and full summaries of books on Project Gutenberg based on web-scraped educational website.
\citep{chen2022summscreen} consists of television show transcripts and episode summaries based on web-scraped fan-written summaries.
The SCROLLS benchmark \citep{shaham2022scrolls} and the MuLD benchmark \citep{hudson2022muld} consist of multiple natural language tasks with long inputs, including long input summarization.
The SQuALITY dataset \citep{wang2022squality} consists of question-focused summaries of Project Gutenberg stories, where annotators write summaries based on different questions that cover different aspects of the same story.

\paragraph{Efficient Transformers}
Many efficient Transformer variants have been introduced in recent years \citep{tay2020efficient}, and we discuss here the works more relevant to this manuscript.
\citep{beltagy2020longformer} use global tokens as well as a sliding window local attention, implemented using custom CUDA kernels.
The ETC model \citep{ainslie2020etc} uses both global tokens and block-wise sliding window local attention, although the global attention is incorporated based on the first few tokens of a sequence, rather than separately learned global tokens.
\citet{zaheer2020bigbird} extend ETC by adding random attention blocks, but we found that this significantly increases code complexity and computational cost.
\citet{guo2021longt5} similarly extend ETC's block-wise sliding window attention, but computes transient ``global token'' representations by pooling over blocks of tokens.
\citet{pang2022topdown} propose to augment the Longformer encoder-decoder with additional pooling layers to improve long-sequence summarization performance.
\citet{ivgi2022sled} propose an alternative approach to sparse attention via encoding overlapping chunks and fusing information across chunks int he decoder.
We highlight that while the final Global-Local model architecture that we settle on shares similarity with several other proposed efficient Transformer architectures, our key contribution lies in our extensive ablation study that identifies architectural tweaks that improve and, just as importantly, do not improve downstream performance.

Among the listed model architectures for long input summarization, LongT5 \citep{guo2021longt5} is the most similar to \largemodel{}, sharing a similar encoder-decoder architecture, a similar training objective in generating masked sentences, and a mix of local attention and global information sharing for the encoder.
We briefly highlight the key differences between the two models.
Firstly, LongT5 trains from scratch on long sequences, whereas we initialize our model weights with PEGASUS weights (which is trained on short sequences) before doing additional pretraining on long input sequences.
This significantly reduces the overall pretraining cost, as short sequence pretraining and be performed much more economically.
LongT5 also uses the T5 relative position biases whereas \largemodel{} uses sinusoidal position embeddings--as shown in Section~\ref{experiments_position_encoding}, T5 relative position biases perform slightly better but are significantly slower.
The efficient encoder architecture between the two models is also different: LongT5 uses a transient global representations based on pooling chunks of tokens, whereas \largemodel{} uses learned global token embeddings.
LongT5 also uses a sliding window local attention based on ETC \citep{ainslie2020etc}, whereas we use a simpler block-local attention with staggered blocks.
Lastly, the largest LongT5 model is 3B parameters, more than 5$\times$ the size of \largemodel{}.

More broadly, \citet{tay2021long} compare a variety of efficient Transformer architectures on a set of tasks designed to probe long-sequence processing capability, evaluating the different models on both performance as well as computation requirements.
\citet{tay2022scaling} further evaluate the scaling properties of novel Transformer architectures, finding that deviating from full attention tends to hurt downstream performance.
\citet{xiong-etal-2022-simple} showed that simple local attention variants can be highly competitive with more complex sparse attention schemes, consistent with our findings.

\section{Conclusion}

In this work, we investigate a range of proposed improvements to allow Transformer models to effectively and economically handle long inputs in text summarization tasks.
Through extensive ablation experiments, we find a simple but effective recipe for extending short input Transformers to tackle long-input summarization.
Based on our findings, we introduce \largemodel{}, an extended version of PEGASUS with a modified architecture and additional long-sequence pretraining.
We show that \largemodel{} sets the state of the art on two long input summarization tasks (GovReport and PubMed) and performs competitively on many others, even despite being much smaller than some compared models.
Our findings can also be applied to extending models to handle long input sequences in other domains beyond summarization, both for pretraining long input models from scratch as well as extending already pretrained short sequence models.

\bibliography{custom}

\newpage

\appendix

\label{appendix:finetuning_hp}

\begin{table*}[th]
\centering
\small
\begin{tabular}{l \rougecolumns \rougecolumns \rougecolumns \rougecolumns}
    \toprule
    & \multicolumn{4}{c}{arXiv} 
    & \multicolumn{4}{c}{GovReport}
    \\
    \cmidrule(lr){2-5}\cmidrule(lr){6-9}
    Position Encoding
    \rougecolumnnames
    \rougecolumnnames
    \\ \midrule
    Factor=10000
        & 48.1 & 20.4 & 28.6 & 30.4
        & 60.9 & 29.3 & 30.8 & 38.0
    \\
    Factor=50000
        & 48.1 & 20.4 & 28.6 & 30.4
        & 61.4 & 29.5 & 30.9 & 38.3
    \\
    \bottomrule
\end{tabular}%
\caption{
  Comparison of different scaling constants in sinusoidal position encodings.
}
\label{tab:table_a1_position_encoding_scale}
\end{table*}

\section{Fine-tuning Hyperparameters}

The hyperparameters for fine-tuning models are shown in Table~\ref{tab:table_a2_finetuning_hyperparameters}. 

\begin{table*}[th]
\centering
\small
\begin{tabular}{l ccccccc}
    \toprule
    Dataset
    & \multirow{2}{*}{\shortstack{Batch \\ Size}}
    & \multirow{2}{*}{\shortstack{Learning \\ Rate}}
    & \multirow{2}{*}{\shortstack{Num \\ Steps}}
    & \multirow{2}{*}{\shortstack{Max Input \\ Tokens}}
    & \multirow{2}{*}{\shortstack{Max Output \\ Tokens}}
    & \multirow{2}{*}{\shortstack{Beam \\ Size}}
    & \multirow{2}{*}{\shortstack{Beam \\ Alpha}}
    \ \\
    \\ \midrule
    \multicolumn{8}{c}{\smallmodel}
    \\ \midrule
    XSum 
    & 64 & 8e-4 & 97.5K & 1024 & 128 & 4 &  0.8
    \\
    CNN/DailyMail
    & 64 & 8e-4 & 410K & 1024 & 128 & 4 &  0.8
    \\
    arXiv
    & 64 & 8e-4 & 92.5K & 16384 & 256 & 1 &  1
    \\
    Big Patent
    & 64 & 8e-4 & 272.5K & 16384 & 256 & 1 &  1
    \\
    PubMed
    & 64 & 8e-4 & 85K & 8096 & 256 & 1 &  1
    \\
    GovReport
    & 64 & 8e-4 & 40K & 12288 & 1024 & 2 &  1
    \\
    SummScreen
    & 64 & 8e-4 & 90K & 16384 & 256 & 1 &  1
    \\
    QMSum
    & 64 & 8e-4 & 7.5K & 16384 & 256 & 1 &  1
    \\
    \\ \midrule
    \multicolumn{8}{c}{\largemodel}
    \\ \midrule
    XSum 
    & 64 & 8e-4 & 5k & 1024 & 128 & 4 &  0.8
    \\
    CNN/DailyMail
    & 64 & 8e-4 & 7.5k & 1024 & 128 & 4 &  0.8
    \\
    arXiv
    & 64 & 8e-4 & 85k & 16384 & 256 & 1 &  1
    \\
    Big Patent
    & 64 & 8e-4 & 390k & 12192 & 256 & 1 &  1
    \\
    PubMed
    & 64 & 8e-4 & 47.5k & 12192 & 256 & 1 &  1
    \\
    GovReport
    & 64 & 8e-4 & 75K & 12288 & 1024 & 1 &  1
    \\
    SummScreen
    & 64 & 8e-4 & 40K & 12192 & 256 & 1 &  1
    \\
    QMSum
    & 64 & 8e-4 & 35K & 12192 & 256 & 1 &  1
    \\
    \bottomrule
\end{tabular}%
\caption{
  Hyperparameters for fine-tuning models
}
\label{tab:table_a2_finetuning_hyperparameters}
\end{table*}

\section{Engineering Details}

The original PEGASUS model was trained using a codebase based on TensorFlow.
The experiments in this paper were run using a new codebase written with JAX \citep{jax2018github} and Flax \citep{flax2020github}.
\smallmodel and \largemodel were trained by converting the weights from the TensorFlow checkpoint to a Flax checkpoint format, and then continuing with long input training.

\end{document}